\documentclass[10pt,twocolumn,letterpaper]{article}

\usepackage{iccv}
\usepackage{times}
\usepackage{epsfig}
\usepackage{graphicx}
\usepackage{amsmath}
\usepackage{amssymb}

\usepackage{multirow}
\usepackage{tabularx}
\usepackage{booktabs}

\usepackage[accsupp]{axessibility}  

\usepackage[font=footnotesize,labelfont=bf]{caption}
\usepackage[belowskip=-5pt,aboveskip=3pt]{caption} 

\usepackage{dsfont}
\usepackage{url}
\usepackage{color}
\usepackage{amsthm}
\usepackage[export]{adjustbox}
\usepackage{comment}

\usepackage[utf8]{inputenc} 
\usepackage[T1]{fontenc}    
\usepackage{url}            
\usepackage{amsfonts}       
\usepackage{nicefrac}       
\usepackage{microtype}      
\usepackage{enumitem}
\usepackage{blindtext}
\usepackage{sidecap}

\iccvfinalcopy

\definecolor{citecolor}{RGB}{65,105,225}

\usepackage[pagebackref=true,breaklinks=true,letterpaper=true,colorlinks,bookmarks=false,citecolor=citecolor]{hyperref}

\usepackage{dsfont}
\definecolor{dg}{rgb}{0,0.694,0.298}
\definecolor{purple}{rgb}{0.4,0.176,0.569}
\definecolor{royalblue}{RGB}{65,105,225}
\usepackage{pifont}
\newcommand{\figref}[1]{Fig.~\ref{#1}}

\newcommand{\secref}[1]{Sec.~\ref{#1}}
\newcommand{\tableref}[1]{Table~\ref{#1}}

\makeatletter
\DeclareRobustCommand\onedot{\futurelet\@let@token\@onedot}
\def\@onedot{\ifx\@let@token.\else.\null\fi\xspace}
\def\eg{\emph{e.g}\onedot} 
\def\ie{\emph{i.e}\onedot} 
 
\def\etc{\emph{etc}\onedot} 
\def\wrt{w.r.t\onedot} 

\makeatother

\definecolor{americanrose}{rgb}{1.0, 0.01, 0.24}

\newcommand{\topone}[1]{\textbf{\textcolor{red}{#1}}}

\usepackage[capitalize]{cleveref}
\crefname{section}{Sec.}{Secs.}
\Crefname{section}{Section}{Sections}
\Crefname{table}{Table}{Tables}
\crefname{table}{Tab.}{Tabs.}

\def\iccvPaperID{4238} 

\ificcvfinal\fi

\begin{document}

\title{Leveraging Inpainting for Single-Image Shadow Removal}

\author{Xiaoguang Li\textsuperscript{1}\thanks{Xiaoguang Li and Qing Guo are co-first authors and contribute equally.}~, 
~Qing Guo\textsuperscript{2$*$}\thanks{Qing Guo (\href{mailto:guo\_qing@cfar.a-star.edu.sg}{guo\_qing@cfar.a-star.edu.sg}) is the corresponding author.},
~Rabab Abdelfattah\textsuperscript{1},
~Di Lin\textsuperscript{3},\\
Wei Feng\textsuperscript{3},
~Ivor Tsang\textsuperscript{2},
~Song Wang\textsuperscript{1}
\\~\\
\textsuperscript{1}University of South Carolina, USA,
\textsuperscript{2} IHPC and CFAR, Agency for Science, \\ Technology and Research, Singapore, 
\textsuperscript{3} Tianjin University, China
}

\ificcvfinal\thispagestyle{empty}\fi


\twocolumn[{%
\renewcommand\twocolumn[1][]{#1}%
\maketitle
\begin{center}
\centering
\includegraphics[width=1.0\textwidth]{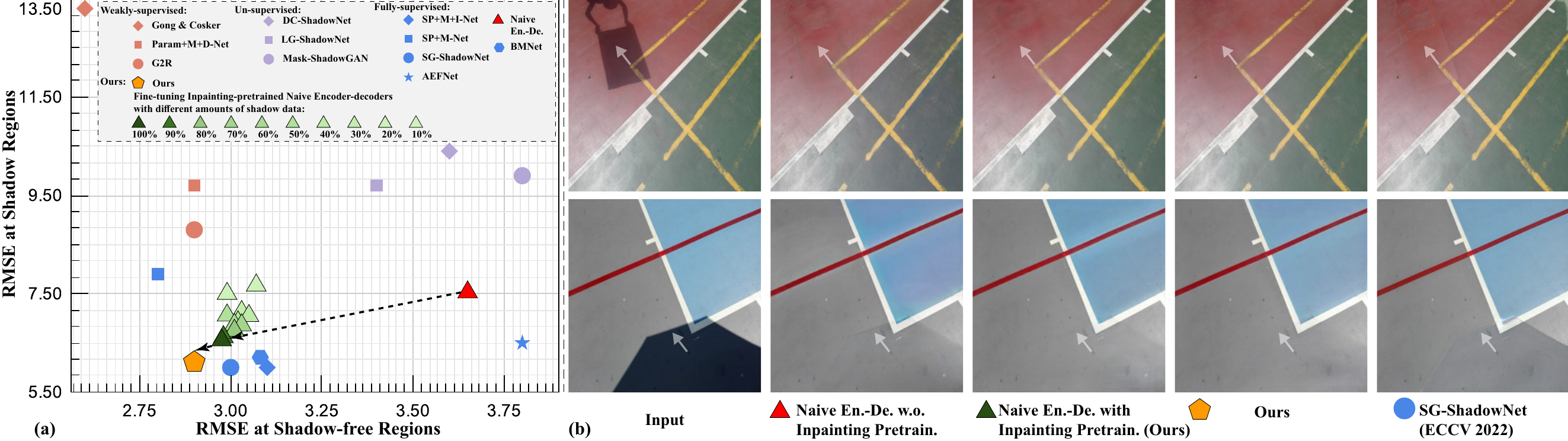}
\captionof{figure}{ (a) Comparison with SOTA methods on ISTD+ dataset\cite{le2019shadow} at shadow (Y-axis) and shadow-free (X-axis) regions. We also present the results of a naive encoder-decoder network (Naive En.-De.) trained with different strategies: (1) training on inpainting data and fine-tuning on different amounts of shadow data; (2) only training on all shadow data. (b) Displays two examples of four methods and we use white arrows to highlight the main advantages.}
\label{fig:mot}
\end{center}
}]


\begin{abstract}
Fully-supervised shadow removal methods achieve the best restoration qualities on public datasets but still generate some shadow remnants.
One of the reasons is the lack of large-scale shadow \& shadow-free image pairs.
Unsupervised methods can alleviate the issue but their restoration qualities are much lower than those of fully-supervised methods.
In this work, we find that pretraining shadow removal networks on the image inpainting dataset can reduce the shadow remnants significantly: a naive encoder-decoder network gets competitive restoration quality \wrt the state-of-the-art methods via only 10\% shadow \& shadow-free image pairs.
%
After analyzing networks with/without inpainting pretraining via the information stored in the weight (IIW), we find that inpainting pretraining improves restoration quality in non-shadow regions and enhances the generalization ability of networks significantly. Additionally, shadow removal fine-tuning enables networks to fill in the details of shadow regions. 
Inspired by these observations we formulate shadow removal as an adaptive fusion task that takes advantage of both shadow removal and image inpainting. 
Specifically, we develop an adaptive fusion network consisting of two encoders, an adaptive fusion block, and a decoder. The two encoders are responsible for extracting the features from the shadow image and the shadow-masked image respectively. The adaptive fusion block is responsible for combining these features in an adaptive manner. 
Finally, the decoder converts the adaptive fused features to the desired shadow-free result. The extensive experiments show that our method empowered with inpainting outperforms all state-of-the-art methods.
We have realized codes and models in \url{https://github.com/tsingqguo/inpaint4shadow}
\end{abstract}

\section{Introduction}
\label{sec:intro}

Shadows in images are formed when some objects block the light source, which not only reduces the image quality but also affects the subsequent intelligent tasks like visual object detection \cite{nadimi2004physical}\cite{fu2021deep}, objects tracking \cite{sanin2010improved},  face recognition \cite{zhang2018improving}, face landmark detection \cite{fu2021benchmarking}, \etc. 
Single-image shadow removal is to map the shadow regions to their shadow-free counterparts, which can enhance the visual quality and benefit the intelligent tasks.

In recent years, with the development of advanced learning algorithms and deep architectures (\ie, GAN \cite{goodfellow2020generative} and CycleGAN \cite{zhu2017unpaired}), deep learning-based shadow removal methods \cite{le2020from,fu2021auto,wan2022sg-shadow} have achieved significant progress and obtained top restoration qualities on public datasets. 
However, these methods may still lead to some shadow remnants. As the two cases shown in \figref{fig:mot}, even the state-of-the-art method SG-ShadowNet \cite{wan2022sg-shadow} generates obviously inconsistent colors across shadow boundaries. 
One of the reasons is the lack of large-scale shadow~\&~shadow-free image pairs.
The commonly-used ISTD+ dataset\cite{le2019shadow}  only contains 1330 pairs of shadow and shadow-free images for training. 
To alleviate the requirements for the amounts of image pairs, researchers also develop weakly-supervised \cite{gong2016interactive,le2020from,liu2021from} and unsupervised shadow removal methods \cite{jin2021dc,liu2021shadow,hu2019maskshadowgan,liu2021from}. 
However, all these methods still have a large gap to the fully-supervised methods (See the results in \figref{fig:mot}~(a)). 

Meanwhile, several works \cite{he2022masked,xie2022simmim} have demonstrated that pretraining networks to predict masked patches from unmasked patches on a large-scale dataset can enhance the fully-supervised training on another small-scale dataset significantly. 
Actually, the task predicting masked patches is a special image inpainting task \cite{li2022misf} that aims to fill missing pixels. 
Inspired by these works, we seek to leverage the image inpainting task for high-quality shadow removal. 
Specifically, given a deep network, we first train it on the image inpainting dataset where we can mask all clean images randomly and get large-scale clean~\&~masked image pairs for free.
Then, we finetune the network on the shadow removal dataset with limited shadow~\&~shadow-free image pairs.
As shown in \figref{fig:mot}, a naive encoder-decoder network pretrained on the inpainting dataset and fine-tuned on the 10\% shadow data can achieve competitive restoration qualities (\ie RMSE) at the shadow regions \wrt the state-of-the-art methods (\eg, SP+M-Net \cite{le2019shadow}).
Moreover, when we compare the networks with and without inpainting pretraining in the \figref{fig:mot}~(b), we see that inpainting pretraining can eliminate the shadow remnants effectively.

%
%
We further analyze the features of the networks with/without inpainting pretraining and see that inpainting pretraining not only improves restoration quality in non-shadow regions but also enhances the generalization ability of networks. This improvement in generalization ability can be proved by using the PAC-Bayes theorem to measure the amount of information stored in the network's weights (IIW) \cite{wang2021pac}. Furthermore, fine-tuning the networks for shadow removal enables them to fill in the details of shadow regions, thereby improving the overall restoration quality of the images.
To utilize the respective advantages of shadow removal and image inpainting, we formulate shadow removal as an adaptive fusion task of shadow removal and image inpainting where make the shadow removal and image inpainting focus on the shadow and non-shadow regions respectively. 
%
%
To address this task, we propose an adaptive fusion network which consisting of two encoders, an adaptive fusion block, and a decoder. The two encoders are responsible for extracting the features from the shadow image and the shadow-masked image respectively. The adaptive fusion block is responsible for combining the extracted features in an adaptive manner and the decoder converts the adaptively fused features into the desired shadow-free result. 
Our final method outperforms all state-of-the-art methods on the public datasets (See \figref{fig:mot}(a)) and reduces the shadow remnants significantly.


\section{Related Work}
\label{sec:formatting}

\subsection{General Shadow Removal Methods}

To reconstruct the shadow-free image from its shadow counterpart, previous methods\cite{finlayson2005removal}\cite{finlayson2009entropy}\cite{guo2012paired}\cite{yang2012shadow}\cite{xiao2013fast}\cite{gryka2015learning}\cite{ma2016appearance} focus more on exploiting the priors information. \cite{guo2012paired} utilize illumination difference to distinguish the shadow and shadow-free regions, then recover the shadow-free image by a lighting model. \cite{finlayson2005removal} \cite{finlayson2009entropy} rebuild the shadow-free image based on the gradient information. \cite{gryka2015learning} proposes to learn a patch mapping function to perform shadow removal.

Recently, deep learning-based methods have achieved excellent performance in shadow removal. \cite{le2019shadow} decomposes the shadow image as its shadow-free counterpart and a set of shadow parameters and uses illumination transformation to remove the shadow. \cite{zhu2022bijective} claims that the networks of shadow removal and generation  can mutually promote, then proposes a unified framework to perform shadow removal and generation together. \cite{chen2021canet}\cite{wan2022sg-shadow} design a feature transformation network to transform the contextual information from non-shadow regions to shadow regions. \cite{fu2021auto} reformulates shadow removal as an exposure fusion problem, utilizing the deep neuron network to predict the parameters of exposure. Although achieving a remarkable result, the above supervised-based methods require a large amount of paired shadow \& shadow-free images to train the network. To alleviate this limitation, the unsupervised methods \cite{jin2021dc}\cite{liu2021shadow}\cite{hu2019maskshadowgan}\cite{liu2021from} employ the generative adversarial network (GAN) to train the network with large numbers of unpaired shadow \& shadow-free images. However, the restoration results of unsupervised methods are not satisfied. In this work, we formulate shadow removal as an adaptive fusion task and propose a novel adaptive fusion network for shadow removal.


\subsection{Predictive Filtering Technique}

The predictive filtering technique has been widely used for image processing\cite{bako2017kernel}\cite{mildenhall2018burst}\cite{cheng2020adversarial}\cite{guo2020watch}\cite{zhai2020s}\cite{lin2022generative}. Different from the traditional convolutional neural network (CNN) which shares the same kernel along the spacial level for each convolution layer, the predictive filtering technique predicts unique kernel for each location by using a deep neural network which makes it can utilize the neighbors' information explicitly. \cite{guo2021efficientderain} use the predictive filtering on image level for deraining. \cite{guo2021jpgnet} employ it to improve the performance of the generative adversarial network (GAN) based inpainting methods. \cite{li2022misf} extend the image level filtering into the feature level and design an end-to-end image inpainting pipeline that can restore the images with large missing regions. 
In this work, we exploit the mechanism of the predictive filtering technique to predict the adaptive weights in an element-wise manner which can be used to take information adaptively from the extracted features.

\section{Leveraging Inpainting-pretrained Networks for Shadow Removal}
\label{sec:section3}

In this section, we first train a naive encoder-decoder network through an inpainting dataset (See \secref{subsec:train_inpaint}). 
%
%
Then, we train the same network with paired shadow~\&~shadow-free images (See \secref{subsec:finetune_inpaint}) based on the inpainting-pretrained network and randomly initialized network, respectively. 
We conduct extensive empirical study and analysis in \secref{subsec:observations}.

\subsection{Training Network for Image Inpainting}
\label{subsec:train_inpaint}

Given a corrupted image $\mathbf{I}$ where some regions are cropped and indicated by a mask $\mathbf{M}$, an image inpainting network is to fill the missing pixels and generate an image that is desired to be identical to the ground truth image (\ie, $\mathbf{I}^*$).
To train the network, we obtain the corrupted image $\mathbf{I}$ by cropping the ground truth image $\mathbf{I}^*$ according to the mask, and then the image pair $(\mathbf{I},\mathbf{I}^*)$ forms a training example.

Here, we use the following setups for training an image inpainting network:
\ding{182} Network architecture. Instead of using advanced inpainting networks (\eg, MISF \cite{li2022misf} and CICM \cite{feng2022cross}), we employ a naive encoder-decoder network that contains 11 convolution layers for the encoder and 3 transpose convolution layers for the decoder. 
%
%
By doing this, we can avoid the effects of some advanced designs on the observations.
\ding{183} Dataset and masks. We train the encoder-decoder network on the \textit{Places2 dataset}\cite{zhou2017places} that is widely used in the inpainting field. We randomly sample masks from a third-party mask dataset \cite{liu2018image} and use them to crop the clean images.
\ding{184} Loss functions. We follow the designs of \cite{li2022misf} and use the GAN loss, perceptual loss, style loss, and $L_1$ loss for training. 
%
With the above setups, we can train a naive encoder-decoder network for image inpainting. 

\subsection{Training Network for Shadow Removal}
\label{subsec:finetune_inpaint}

Given a shadow image $\mathbf{I}$ and a shadow mask $\mathbf{M}$, a shadow removal network is to map the shadow regions indicated by the mask to their shadow-free counterparts and is desired to produce a clean image $\mathbf{I}^{*}$. 
In contrast to the image inpainting task, we can hardly collect numerous paired images (\ie, $(\mathbf{I},\mathbf{I}^*)$) for training the shadow removal network since the shadow image cannot be naively simulated via the ground truth image $\mathbf{I}^*$ and the two images should be captured under the same scene and the same environmental factors. Even the widely-used ISTD+ dataset\cite{le2019shadow} only contains 1,330 pairs of shadow and shadow-free images for training.

Here, we use the following setups for training a shadow removal network: \ding{182} Network architecture. We employ the same encoder-decoder network in \secref{subsec:train_inpaint}.\ding{183} Dataset and masks. We use the ISTD+ dataset\cite{le2019shadow}  for network training and testing where the shadow mask for each image is given. \ding{184} Loss functions. We only use $L_1$ loss for training.
%
Then, we set up two training strategies: \textit{First}, we randomly initialize the network's weights and train the network on the ISTD+ dataset\cite{le2019shadow}. \textit{Second}, we use the inpainting-pretrained network as the initialization and fine-tuning the network via the ISTD+ dataset\cite{le2019shadow}.
Based on the above setups, we can study how inpainting affects the shadow removal network training.

%
\begin{figure}[t]
\centering
\includegraphics[width=1.0\linewidth]{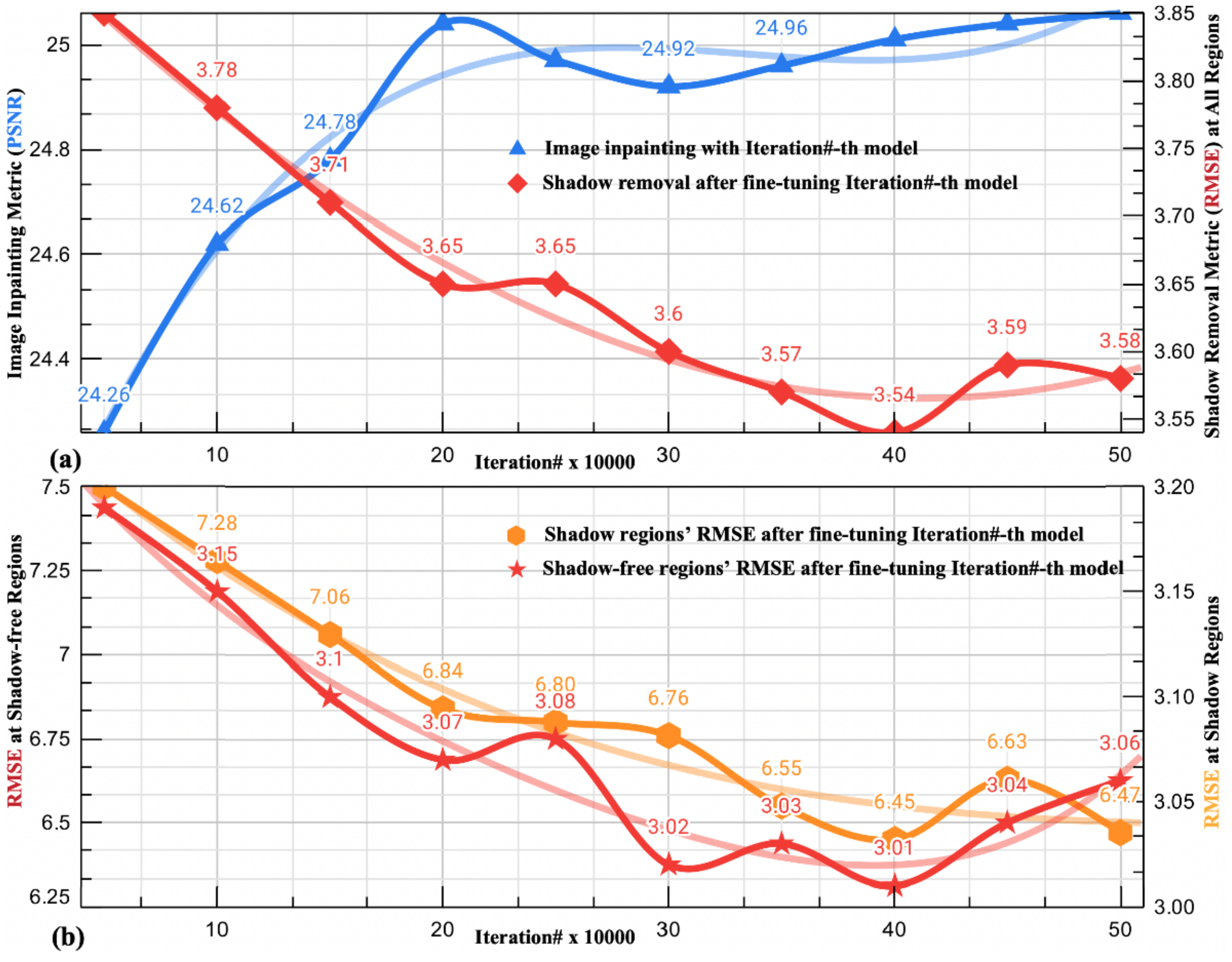}
\caption{
We first train the naive encoder-decoder network on the Place2 dataset and save the networks every $5\times 10^{5}$ iterations, and then fine-tune each model with the shadow removal dataset (\ie, ISTD+\cite{le2019shadow}). The blue line in (a) presents the inpainting results (\ie, PSNR) of different networks on the Places2 dataset, and the red line shows shadow removal results (\ie, root
mean square error (RMSE)) of fine-tuned networks on both shadow and shadow-free regions. (b) presents the shadow removal results (\ie, RMSE) on the shadow regions and shadow-free regions, respectively.
}
\label{fig:empirical}
\vspace{-15pt}
\end{figure}
%

\subsection{Empirical Study and Analysis}
\label{subsec:observations}

{\bf Fine-tuning inpainting-pretrained networks for shadow removal.} 
Following the setups in \secref{subsec:train_inpaint}, we first train an encoder-decoder network for image inpainting and store the intermediate models every $5\times 10^{5}$ iterations. Then, we fine-tune these intermediate models on the shadow removal dataset (See \secref{subsec:finetune_inpaint}). 
We show the inpainting results (\ie, PSNR) of the intermediate models and the corresponding shadow removal performance (\ie, RMSE) after fine-tuning in \figref{fig:mot}(a) and \figref{fig:empirical}. 
We see that: \ding{182} All shadow removal networks (\ie, green triangles in \figref{fig:mot}(a)) that are fine-tuned on the inpainting-pretrained models achieve much lower RMSEs at both shadow and shadow-free regions than the network trained on the randomly initialized model (\ie, red triangle in \figref{fig:mot}(a)).
This demonstrates that inpainting pretraining indeed helps shadow removal task get much higher restoration quality in both shadow and shadow-free regions. 
\ding{183} As the iteration becomes larger, the inpainting capability of intermediate models increases (\ie, the PSNR on the inpainting dataset increases) and the corresponding fine-tuned shadow removal networks get lower RMSEs at the shadow and shadow-free regions. This observation demonstrates that the shadow removal performance is related to the inpainting capability of pretrained models, directly.

{\bf Fine-tuning inpainting-pretrained networks with different amounts of shadow \& shadow-free image pairs.} 
With an adequately inpainting-pretrianed network, 
we use different amounts of shadow \& shadow-free image pairs that are randomly selected from the ISTD+ \cite{le2019shadow} to fine-tune the network, respectively. As shown in \figref{fig:mot}~(a), we see that: \ding{182} With only 10\% ISTD+ \cite{le2019shadow}, the fine-tuned encoder-decoder network achieves lower RMSE at shadow-free regions than the state-of-the-art method AEFNet \cite{fu2021auto} (\ie, 3.07 vs. 3.80) and lower RMSE on shadow regions than the SP+M-Net method (\ie, 7.67 vs. 7.9). Both methods are trained with all examples in ISTD+ dataset\cite{le2019shadow}. \ding{183} Inpainting-pretrained networks are fine-tuned with different amounts of image pairs and have similar RMSEs at both shadow and shadow-free regions. 
\ding{184} The network fine-tuned with 100\% image pairs outperforms three recent works including SG-ShadowNet \cite{wan2022sg-shadow}, SP+M+I-Net \cite{le2021physics}, and AEFNet \cite{fu2021auto} at the shadow-free regions, which demonstrates that inpainting pretraining benefits the shadow-free preservation.

%
\begin{figure}[t]
\centering
\includegraphics[width=1.0\linewidth]{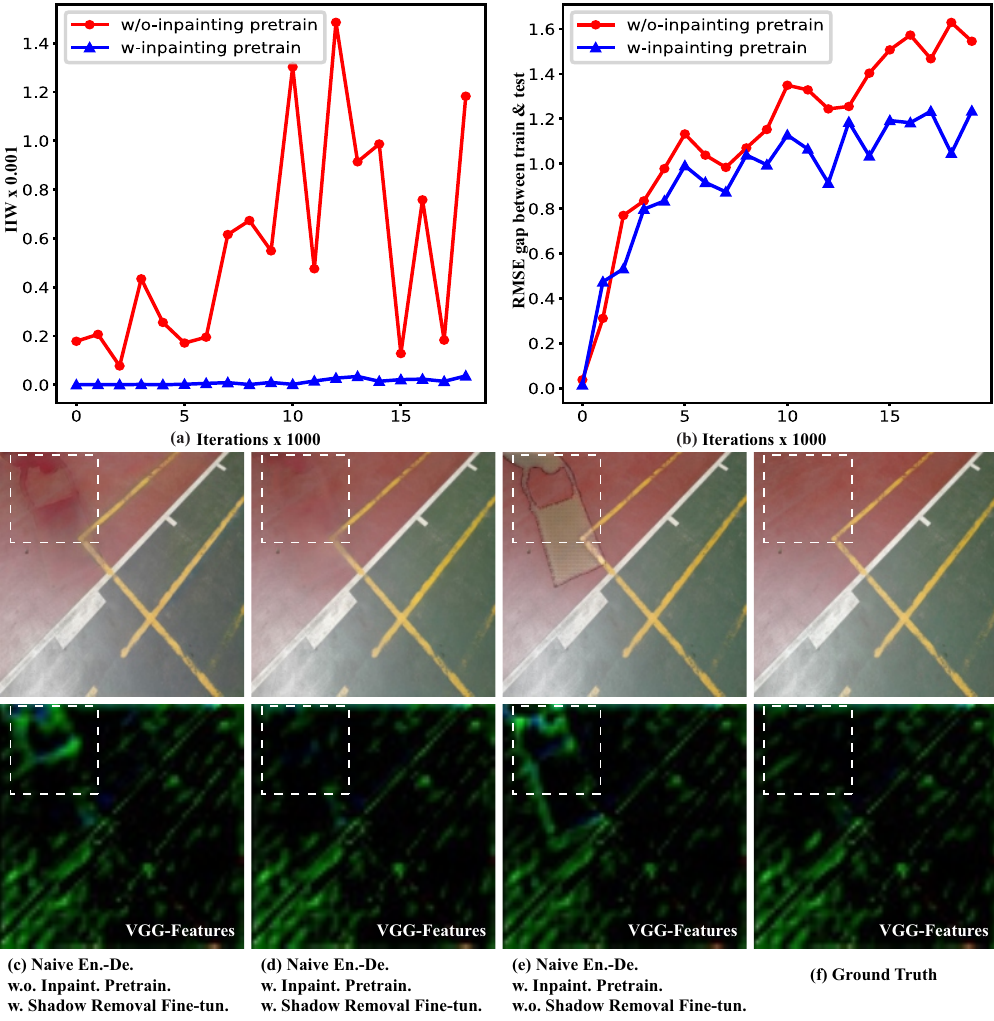}
\caption{
(a) Analysis with/without inpainting pretraining via IIW. %
(b) RMSE gap between the training and testing datasets.
(c)-(f) Visualized comparison with/without inpainting pretraining and shadow removal fine-tuning.
}
\label{fig:features}
\vspace{-15pt}
\end{figure}
%


{\bf Analysis with/without inpainting pretraining via information stored in the weight (IIW) \cite{wang2021pac}.}
To further understand the benefits of using inpainting pretraining for shadow removal, we fine-tune a naive encoder-decoder network both with and without inpainting pretraining on the ISTD+ dataset and evaluate the network's performance every 1000 iterations. To assess the network's performance, we monitor two key metrics.
First, we use the PAC-Bayes theorem to measure the amount of information stored in the network's weights (IIW) \cite{wang2021pac}, which is a promising indicator of a deep neural network's generalization ability. As shown in \figref{fig:features} (a), with inpainting pretraining, the network's generalization capability improved significantly.
In addition to the IIW, we also calculated the Root Mean Squared Error (RMSE) gap, which is the difference in RMSE values obtained by evaluating the network on the training and testing datasets. As shown in \figref{fig:features} (b), the RMSE gap was smaller with inpainting pretraining than without it which can prove that inpainting pretraining can decrease the degree of overfitting of the network.

{\bf Visualized comparison with/without inpainting pretraining and shadow removal fine-tuning.} 
We also analyze the performance of with/without inpainting pretraining in both image and feature level as shown in \figref{fig:features} (c)-(f). Comparing \figref{fig:features}~(c) with (d), we see that the inpainting pretrained network can suppress the shadow patterns (\ie, the white dashed rectangles) at both the semantic level and image level effectively. Comparing \figref{fig:features}~(d) and (e), we see that without the shadow removal fine-tuning, the network cannot recover the colors properly and the shadow patterns become more obvious. We have similar observations at the feature level.

\begin{figure*}[t]
\centering
\includegraphics[width=1.0\linewidth]{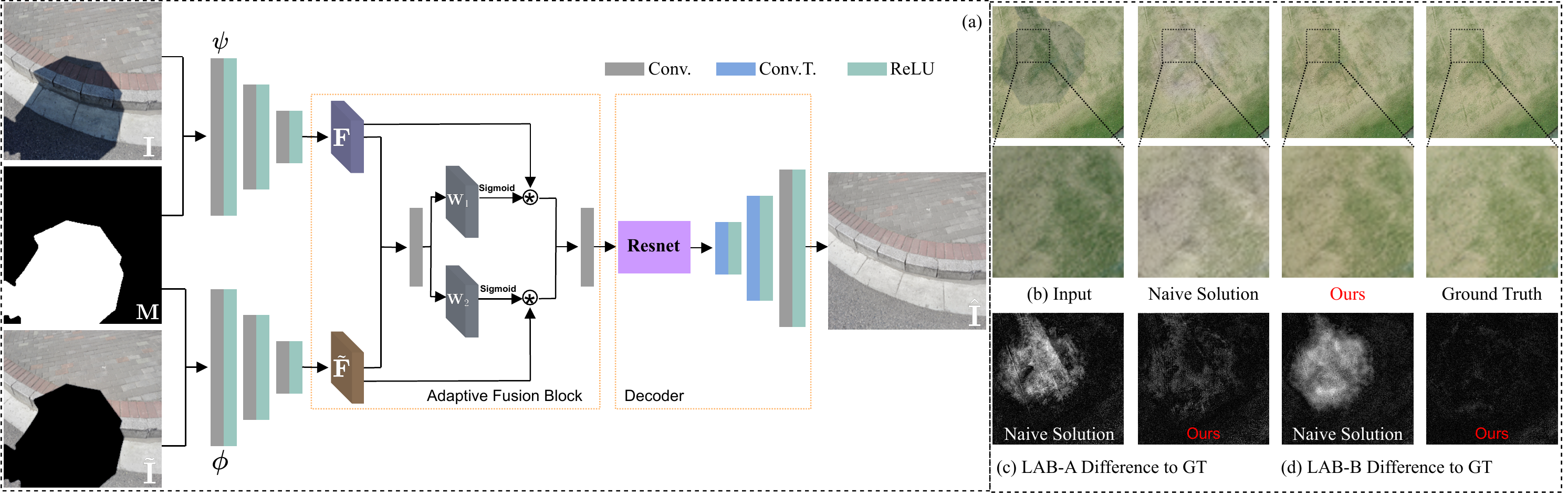}
\caption{
(a) Displays the whole framework of the proposed adaptive fusion network. (b) Shows examples of using a simple solution that input the concatenated shadow image and shadow-masked image into a naive encoder-decoder and the proposed method to handle the shadow image, respectively. We also display the enlarged regions in the second row. We further calculate the color difference to the ground truth of the A channel (\ie, (c)) and B channel (\ie, (d)) in the LAB color space.
}
\label{fig:framework}
\vspace{-15pt}
\end{figure*}
%

\section{Methodology}
\label{sec:method}
In \secref{sec:section3}, we demonstrate the significant benefits of inpainting pretraining for shadow removal. Building upon these insights, in \secref{sec:method}, we explicitly integrate inpainting into shadow removal to fully harness its potential capabilities and formulate the shadow removal as a special adaptive fusion task, as detailed in \secref{subsec:shadowinpaint}.
To this end, we propose an adaptive fusion network in \secref{subsec:shadowfilter} to realize the task in a unified architecture and detail the implementations in \secref{subsec:imp}.
\subsection{Problem Formulation}
\label{subsec:shadowinpaint}

Given a shadow image $\mathbf{I}$ and a binary mask $\mathbf{M}$, we first crop the shadow regions and produce a shadow-masked image $\tilde{\mathbf{I}}=\mathbf{I}\odot(1-\mathbf{M})$. 
Then we aim to reconstruct the shadow-free counterpart based on the shadow image $\mathbf{I}$ and shadow-masked image $\tilde{\mathbf{I}}$.
We denote the whole process as an \textit{adaptive fusion} task.
Intuitively, such a task inherits the advantages of image inpainting and could fill the missing semantic information with similar color distributions as the shadow-free regions (See discussions in \secref{subsec:observations}). 
Besides, the task could recover missing details in the shadow regions since the whole information is contained in the raw shadow image. 
However, the implementation of this task is not straightforward.

A simple solution is to concatenate the shadow-masked image and the shadow image and feed them into the same encoder-decoder network used in the \secref{subsec:train_inpaint}.
We can also perform inpainting pretraining and shadow removal fine-tuning to that network.
We show an example in \figref{fig:framework} (b), (c), and (d).
Clearly, the result of the naive solution still has an obvious color shifting to the ground truth (See the LAB-A/B differences in \figref{fig:framework}~(c) and (d)), which demonstrates that such a naive solution cannot properly take the advantages of both tasks and the color inconsistency still remains.

\subsection{Adaptive Fusion Network}
\label{subsec:shadowfilter}

Instead of concatenating the two images, we propose to handle the shadow image and shadow-masked image via two encoders which are responsible for extracting the feature from the shadow image and shadow-masked image respectively. We show the whole framework in \figref{fig:framework}(a).
we denote the encoder receiving the shadow-masked image as $\phi(\cdot)$.
we can represent the feature $\tilde{\mathbf{F}}\in\mathds{R}^{H_l\times W_l\times C_l}$ extracted from shadow-masked image as
%
\begin{align}\label{eq:inpaint}
   \tilde{\mathbf{F}} = \phi_l(\ldots(\ldots\phi_2(\phi_1([\tilde{\mathbf{I}},\mathbf{M}]))),
\end{align}
%
where $\phi_l(\cdot)$ is the $l$th layer in the encoder.
Meanwhile, we can use another encoder (\ie, $\psi(\cdot)$) to get the  feature of the shadow image 
%
\begin{align}\label{eq:shadow}
   \mathbf{F} = \psi_l(\ldots(\ldots\psi_2(\psi_1([\mathbf{I},\mathbf{M}]))).
\end{align}
%

\begin{figure}[t]
\centering
\includegraphics[width=1.0\linewidth]{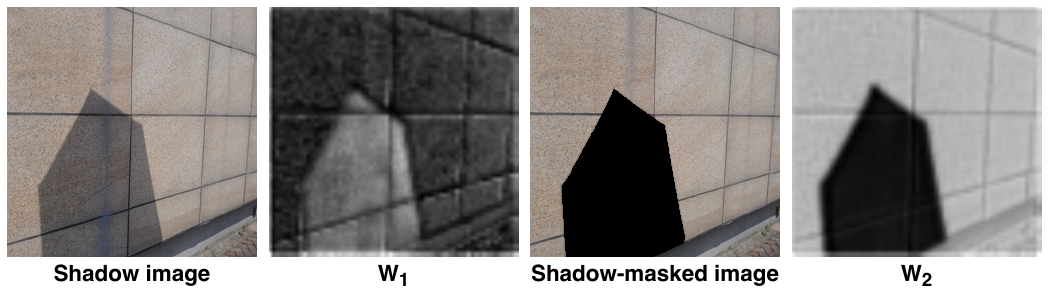}
\vspace{-15pt}
\caption{Visualization of the shadow image, shadow-masked image, and the corresponding weights $\mathbf{W}_1$ and $\mathbf{W}_2$. 
}
\vspace{-15pt}
\label{fig:w1_w2}
\end{figure}

Instead of passing the shadow-masked feature (\ie, $\tilde{\mathbf{F}}$) and shadow feature (\ie $\mathbf{F}$) to the decoder directly, we design an adaptive fusion block to fuse $\tilde{\mathbf{F}}$ and $\mathbf{F}$ in an adaptive manner. Specifically, we input $\tilde{\mathbf{F}}$ and $\mathbf{F}$ into a convolution layer to predict the adaptive weights followed by a sigmoid operation which can be represented as
\begin{align}\label{eq:dynamicfilter}
   [\mathbf{W}_1,\mathbf{W}_2]  = \text{Sigmoid(Conv}_\text{weight}(\tilde{\mathbf{F}}, \mathbf{F})),
\end{align}
where `$\text{Conv}_\text{weight}$' denotes a convolution layer to predict the weights $\mathbf{W}_1$ and $\mathbf{W}_2$. 
With the predicted weights, we can obtain the adaptive fused feature $\hat{\mathbf{F}}$ by
\begin{align}\label{eq:dynamicfilter}
   \hat{\mathbf{F}} = \text{Conv}_\text{fusion}(\tilde{\mathbf{F}}\circledast\mathbf{W}_1\mathbf{+} \mathbf{F}\circledast\mathbf{W}_2),
\end{align}
where `$\circledast$' is the element-wise multiplication operation and `$\text{Conv}_\text{fusion}$' denotes a convolution layer to fuse the weighted features. 
When we conduct the average pooling along the channels of $\mathbf{W}_1$ and $\mathbf{W}_2$ respectively and illustrate the result, as shown in \figref{fig:w1_w2}, we can observe that: in the shadow regions, the adaptive fused feature $\hat{\mathbf{F}}$ take more information from $\mathbf{F}$ which is extracted from the shadow image; in contrast, in the non-shadow regions, it takes more information from $\tilde{\mathbf{F}}$ which is extracted from the shadow-masked image. 
Finally, the adaptive fused feature $\hat{\mathbf{F}}$ is passed into the decoder to get the de-shadowed result.

\subsection{Implementation Details}
\label{subsec:imp}

{\bf Network architectures.}
The encoders $\phi$ and $\psi$ employ the same architecture.
Specifically, the encoders consist of three convolution layers and each convolution layer is followed by a ReLU activation. 
The extracted features by $\phi$ and $\psi$ are passed into the convolution layer $\text{Conv}_\text{weight}$ to predict the weights followed by a sigmoid operation. The predicted weights will be used to adaptive take information from the features (\ie $\mathbf{F}$ and $\tilde{\mathbf{F}}$) extracted from the shadow image and shadow-masked image. $\text{Conv}_\text{weight}$ use a kernel of size 3x3.
Then the weighted feature is fused by another convolution layer $\text{Conv}_\text{fusion}$ with a kernel of size 3x3.
Finally, the fused feature is passed into the decoder. The decoder consists of eight Resnet blocks and three transposition convolution layers. 
Theoretically, we can add the adaptive fusion operation to all layers, which would lead to high computational complexity. 
Here, we equip the adaptive fusion operation to the last layer features of $\phi$ and $\psi$ in order to save computation.
%
%

{\bf Loss functions.}
When training on the inpainting dataset, we follow \cite{li2022misf} and use GAN loss, perceptual loss, Style loss, and $L_1$ loss, which guides the network to fill in missing semantic information and details.
When fine-tuning the network on the shadow dataset, we only use $L_1$ loss to optimize since this stage mainly focuses on restoration fidelity. 
%

{\bf Training details.}
We first pretrain the network 450,000 iterations with batch size 8 on the Places2 dataset, then fine-tune it with 250,000 iterations under the same batch size on the shadow removal datasets. Following \cite{fu2021auto}, the input images are resized to 256x256. We use Adam as the optimizer to optimize the network with a learning rate of 0.00005 and all the experiments are conducted on the Linux server with two NVIDIA Tesla V100 GPUs.

\section{Experiments}

\subsection{Setups}
{\bf Datasets.}
We first train our network on the image inpainting dataset (\ie, Places2 challenge dataset \cite{zhou2017places}) and then fine-tune it on the ISTD+\cite{le2019shadow} and SRD\cite{qu2017deshadownet} datasets, respectively. Finally, we evaluate the trained networks on the ISTD+\cite{le2019shadow} and SRD datasets\cite{qu2017deshadownet}.
The Places2 dataset contains more than millions of images under over 365 scenes. We use the irregular masks in \cite{liu2018image}.
The ISTD+ dataset\cite{le2019shadow} is constituted of 1330 triplets for training and 540 triplets for testing. We follow the previous method and use ground truth masks during the training. For the evaluation, we follow \cite{fu2021auto} using Ostu's algorithm to calculate the difference between the shadow and shadow-free images to get the masks. 
The SRD dataset\cite{qu2017deshadownet} contains 2680 paired shadow and shadow-free images for training and 408 paired shadow and shadow-free images for testing without providing ground truth masks. Therefore, during the training and testing, we use the detected masks provided by DHAN\cite{cun2020towards}.

{\bf Metrics.}
To prove the effectiveness of our method, we follow the previous method \cite{fu2021auto} to calculate the root mean square error (RMSE) between the reconstructed shadow-free images and the corresponding ground truth in the LAB color space. Besides, we also use the peak signal-to-noise ratio (PSNR\cite{johnson2016perceptual}),  structural similarity index (SSIM), and the learned perceptual image patch similarity (LPIPS\cite{zhang2018unreasonable}) to measure the quality of recovered shadow-free images.
To ensure a fair comparison, we reevaluate all baseline methods on the same machine using various metrics.

{\bf Baselines.}
We compare our proposed method with ten state-of-the-art shadow removal methods which includes SP+M-Net\cite{le2019shadow}, DSC\cite{hu2019direction}, DHAN\cite{cun2020towards} , Param+M+D-Net\cite{le2020from}, LG-shadowNet\cite{liu2021shadow}, DC-ShadowNet\cite{jin2021dc}, G2R-ShadowNe\cite{liu2021from}, Fu et al.\cite{fu2021auto}, BMNet\cite{zhu2022bijective}, and SG-ShadowNet\cite{wan2022sg-shadow}.

\subsection{Comparison Results}

\begin{table*}[t]
\footnotesize
\centering
\caption{Comparison results on ISTD+\cite{le2019shadow} and SRD\cite{qu2017deshadownet} datasets.}
\small
{
    \resizebox{0.8\linewidth}{!}{
    {
	\begin{tabular}{c|l|cccc|ccc|ccc}
		
    \toprule
     \multirow{2}{*}{Datasets}
     & \multirow{2}{*}{Method} & \multicolumn{4}{c|}{All} & \multicolumn{3}{c|}{Shadow} & \multicolumn{3}{c}{Non-Shadow}   \\
    &  
    & RMSE$\downarrow$ & PSNR$\uparrow$ & SSIM$\uparrow$ & LPIPS$\downarrow$
    & RMSE$\downarrow$ & PSNR$\uparrow$ & SSIM$\uparrow$
    & RMSE$\downarrow$ & PSNR$\uparrow$ & SSIM$\uparrow$ \\
    
    \midrule
    \multirow{10}{*}{ISTD+}
    
    & SP+M-Net\cite{le2019shadow}
    & 3.610 & 32.33 & 0.9479 & 0.0716
    & 7.205 & 36.16 & 0.9871 
    & 2.913 & 35.84 & 0.9723\\

    & Param+M+D-Net\cite{le2020from} 
    & 4.045	& 30.12	& 0.9420 & 0.0759
    & 9.714 & 33.59 & 0.9850 
    & 2.935 & 34.33 & 0.9723
     \\

    & SynShadow\cite{inoue2020learning}
    & 4.000 & - & - & -
    & 6.900 & - & -
    & 3.400 & - & -
     \\

    & Fu et al.\cite{fu2021auto}
    & 4.278 & 29.43 & 0.8404 & 0.1673
    & 6.583 & 36.41 & 0.9769
    & 3.827 & 31.01 & 0.8755
     \\

    & LG-ShadowNet\cite{liu2021shadow} 
    & 4.402	& 29.20	& 0.9335 & 0.0920
    & 9.709	& 32.65 & 0.9806 
    & 3.363	& 33.36	& 0.9683
     \\

    & DC-ShadowNet\cite{jin2021dc} 
    & 4.781	 & 28.76 & 0.9219 & 0.1112
    & 10.434 & 32.20 & 0.9758
    & 3.674  & 33.21 & 0.9630
     \\
    
    & G2R-ShadowNet\cite{liu2021from} 
    & 3.970 & 30.49	& 0.9330 & 0.0868
    & 8.872 & 34.01 & 0.9770
    & 3.010 & 34.62 & 0.9707
     \\
    
    & BMNet\cite{zhu2022bijective}
    & 3.595 & 32.30 & 0.9551 & 0.0567
    & 6.189 & 37.30 & 0.9899
    & 3.087 & 35.06 & 0.9738
    \\

    & SG-ShadowNet\cite{wan2022sg-shadow}
    & 3.531 & 32.41 & 0.9524 & 0.0594
    & 6.019 & 37.41 & 0.9893
    & 3.044 & 34.95 & 0.9725
     \\

   & \topone{Ours} 
    & \topone{3.398} & \topone{34.14} & \topone{0.9606} & \topone{0.0542} 
    & \topone{5.935} & \topone{38.46} & \topone{0.9894}
    & \topone{2.902} & \topone{37.27} & \topone{0.9772}
    \\

    \midrule
    \multirow{8}{*}{SRD}

    & DSC\cite{hu2019direction} 
    & 5.704	& 29.01	& 0.9044 & 0.1145
    & 8.828	& 34.20 & 0.9702
    & 4.509	& 31.85	& 0.9555
     \\

    & DHAN\cite{cun2020towards} 
    & 4.666	& 30.67	& 0.9278 & 0.0792
    & 7.771 & 37.05 & 0.9818 
    & 3.486	& 32.98	& 0.9591
     \\

    & SynShadow\cite{inoue2020learning}
    & 5.200 & - & - & -
    & 10.90 & - & -
    & 3.600 & - & -
     \\

    & Fu et al.\cite{fu2021auto} 
    & 6.269	& 27.90	& 0.8430 & 0.1820
    & 8.927	& 36.13 & 0.9742
    & 5.259	& 29.43	& 0.8888
     \\

    & DC-ShadowNet\cite{jin2021dc} 
    & 4.893	& 30.75	& 0.9118 & 0.1084
    & 8.103	& 36.68 & 0.9759
    & 3.674	& 33.10	& 0.9540
     \\
    
    & BMNet\cite{zhu2022bijective}
    & 4.240	& 31.88	& 0.9376 & 0.0817
    & 6.982	& 37.41	& 0.9816
    & 3.198	& 35.09	& 0.9676
     \\

    & SG-ShadowNet\cite{wan2022sg-shadow} 
    & 4.297	& 31.31	& 0.9273 & 0.0835
    & 7.564 & 36.55 & 0.9807
    & 3.056	& 34.23	& 0.9611
     \\
    							
    & \topone{Ours} 
    & \topone{3.832} & \topone{33.17} & \topone{0.9398} & \topone{0.0751}
    & \topone{6.094} & \topone{39.33} & \topone{0.9848}
    & \topone{2.973} & \topone{35.61} & \topone{0.9673}
     \\

    \bottomrule

	\end{tabular}
	}
	}
}
\label{tab:comparison_data}
\vspace{-10pt}
\end{table*}


{\bf Quantitative comparison.} 
We first compare our method with the competitors on the ISTD+ dataset\cite{le2019shadow}. As shown in \tableref{tab:comparison_data}, we can observe that: 
\ding{182} for the whole image, our method outperforms all the competitors over each metric \ie RMSE, PSNR, SSIM, and LPIPS. 
Specifically, our method decreases RMSE 5.48\% and LPIPS 4.41\% as well as increases PSNR 5.70\% and SSIM 0.58\% compared with BMNet\cite{zhu2022bijective}. 
%
%
%
%
\ding{183} for the shadow region our method also outperforms other competitors over all the metrics. 
Specifically, compared with LG-shadowNet\cite{liu2021shadow} our method decreases RMSE 38.87\% as well as increases PSNR 17.79\% and SSIM 0.90\% respectively. 
%
%
%
\ding{183} for the non-shadow region, our method still gets the best result.
%
%
Specifically, compared with Fu et al.\cite{fu2021auto} our method decreases RMSE 24.17\%, increases PSNR 20.19\%, and increases SSIM 11.62\% respectively. 

Besides, we further compare with the competitors on the SRD dataset\cite{qu2017deshadownet}. As shown in \tableref{tab:comparison_data}, we see that: 
for the shadow region, non-shadow region, and the whole image, our method outperforms all baselines over the four metrics, \ie, RMSE, PSRN, SSIM, and LPIPS. Compared with DC-ShadowNet\cite{jin2021dc}, 
for the shadow region, our method decreases RMSE 24.80\% as well as increases PSNR 7.22\% and SSIM 0.91\%. 
For the non-shadow region, our method decreases RMSE 19.08\% as well as increases PSNR 7.59\% and SSIM 1.39\%. 
For the whole image, our method decreases RMSE 21.68\% and LPIPS 30.71\% as well as increases PSNR 7.87\% and SSIM 3.07\%.

\begin{figure*}
\centering
\includegraphics[width=1.0\linewidth]{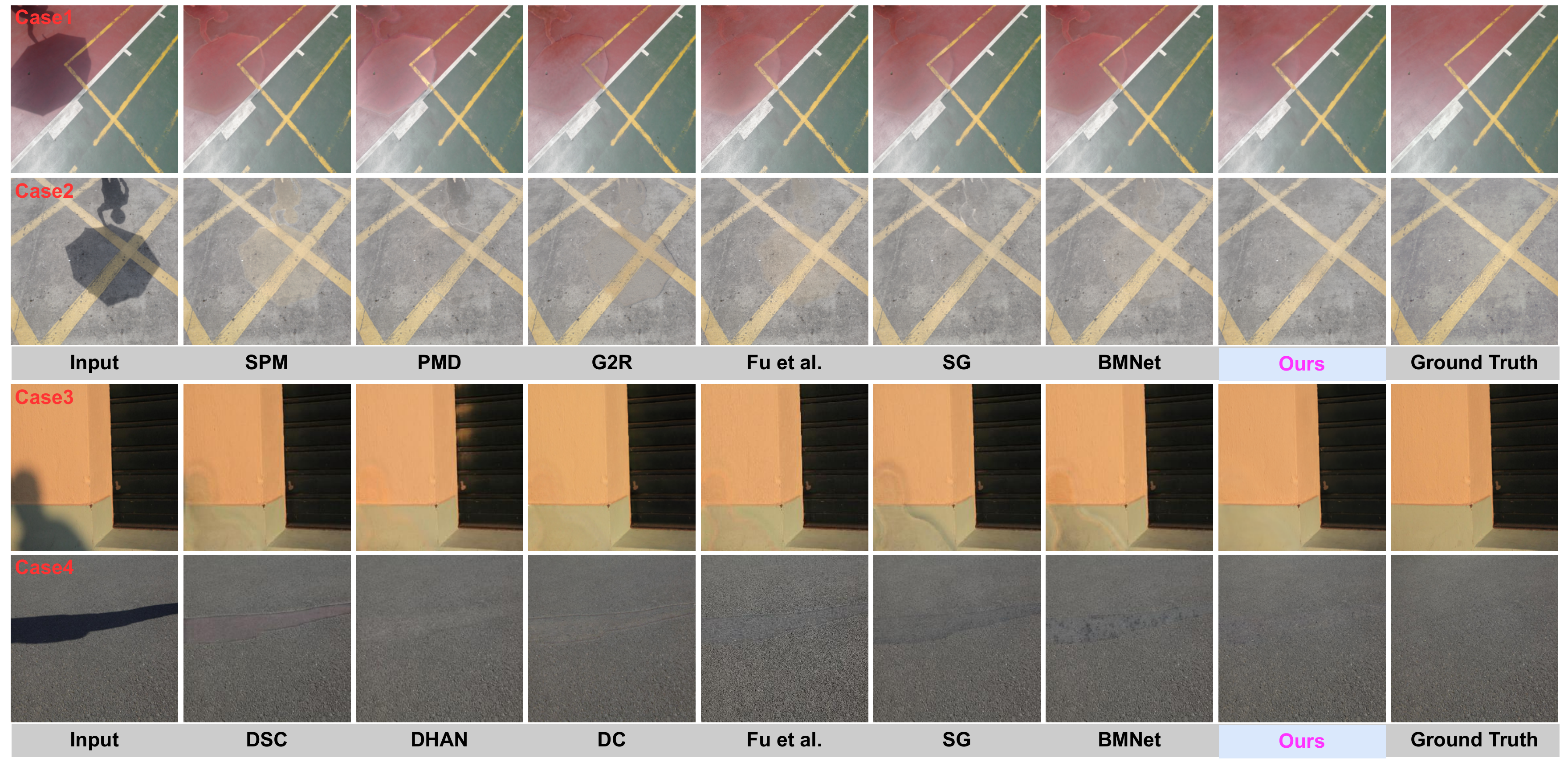}
\vspace{-15pt}
\caption{ Visualization results on ISTD+ dataset\cite{le2019shadow} (first two rows) and SRD dataset\cite{qu2017deshadownet} (last two rows)
}
\label{fig:comparison}
\vspace{-15pt}
\end{figure*}

{\bf Qualitative comparison.} 
We visualize four samples from ISTD+\cite{le2019shadow} and SRD datasets\cite{qu2017deshadownet} in \figref{fig:comparison}. We can observe that \ding{182} our method can generate more natural and realistic color in the shadow regions of the restored result. As shown in Case2 and Case4, the generated floor in the shadow regions by our method is almost identical to the ground truth. Compared with other methods \ie BMNet\cite{zhu2022bijective} and DSC\cite{hu2019direction}, we can find obvious color inconsistencies between the shadow regions and non-shadow regions in their generated result. 
\ding{183} Although without any post process, the generated result by our method is more smooth along the shadow boundary than other methods generated. As shown in Case1, the boundary in our restored result is almost unidentified by human eyes. However, we can observe the boundary clearly in the restored result by other methods. Also in Case3, our restored result is almost identical to the ground truth. But we can find a clear ghost along the shadow boundary in restored results by other methods.

\subsection{Ablation Study}

\begin{table*}[t]
\footnotesize
\centering
\caption{Ablation study on ISTD+ dataset\cite{le2019shadow}.}
\small
{
    \resizebox{0.785\linewidth}{!}{
    {
	\begin{tabular}{l|cccc|ccc|ccc}
		
    \toprule
    \multirow{2}{*}{Method} & \multicolumn{4}{c|}{All} & \multicolumn{3}{c|}{Shadow} & \multicolumn{3}{c}{Non-Shadow}   \\
     
    & RMSE$\downarrow$ & PSNR$\uparrow$ & SSIM$\uparrow$ & LPIPS$\downarrow$
    & RMSE$\downarrow$ & PSNR$\uparrow$ & SSIM$\uparrow$
    & RMSE$\downarrow$ & PSNR$\uparrow$ & SSIM$\uparrow$
    \\
    
    \midrule

    w/o adaptive fusion
    & 3.627 & 33.57 & 0.9573 & 0.0603 
    & 6.928 & 37.68 & 0.9884
    & 2.981 & 36.87 & 0.9756
    \\

    Concatenate inputs
    & 3.581 & 33.58 & 0.9573 & 0.0628 
    & 6.756 & 37.58 & 0.9879
    & 2.959 & 37.03 & 0.9763
    \\

    Encoder-decode w pretraining
    & 3.527	& 33.59	& 0.9589 & 0.0603 
    & 6.377	& 37.66	& 0.9886
    & 2.969	& 36.96	& 0.9769
    \\

    Encoder-decode w/o pretraining
    & 4.216	& 32.75	& 0.9545 & 0.0661 
    & 7.428	& 36.89	& 0.9877
    & 3.588	& 36.09	& 0.9744
    \\

    De-shadow by inpainting
    & 5.992 & 28.62 & 0.8706 & 0.1420 
    & 14.64 & 30.98 & 0.9227
    & 4.298 & 34.58 & 0.9672
    \\

    \topone{Ours} 
    & \topone{3.398} & \topone{34.14} & \topone{0.9606} & \topone{0.0542} 
    & \topone{5.935} & \topone{38.46} & \topone{0.9894}
    & \topone{2.902} & \topone{37.27} & \topone{0.9772}
    \\

    \bottomrule

	\end{tabular}
	}
	}
}
\label{tab:ablation}
\vspace{-16pt}
\end{table*}
To prove the effectiveness of each part of our method, we conduct the following experiments for ablations: \textbf{Exp1}, we remove the adaptive fusion block and combine the extracted features $\tilde{\mathbf{F}}$ and $\mathbf{F}$ directly (See the first row of \tableref{tab:ablation}). We pretrain this variant on the image inpainting dataset and fine-tune that on the shadow dataset. 
\textbf{Exp2}, we concatenate all the inputs and input that into a naive encoder-decoder (See the second row of \tableref{tab:ablation}). We also perform the pretraining and fine-tuning like exp1. 
\textbf{Exp3}, we pretrain a naive encoder-decode network on the image inpainting dataset and fine-tune it on the shadow dataset like Exp1 and Exp2. Meanwhile, we also train another naive encoder-decode directly on the shadow dataset (See the third and fourth row of \tableref{tab:ablation}).
\textbf{Exp4}, we train an inpainting model based on the naive encoder-decoder architecture, then input the shadow-masked image into the inpainting model to perform the shadow removal. 
As shown in \tableref{tab:ablation}, we have the following observations: \ding{182} without the adaptive fusion the performance decreases a lot, \eg the RMSE decreases 14.33\% and 2.65\% in the shadow and non-shadow regions respectively.
\ding{183} Both solutions of concatenating all the inputs and performing the shadow removal by an inpainting model get a worse result in the shadow and non-shadow regions compared with our method. 
Specifically, concatenating all the input decreases the RMSE 12.15\% in the shadow regions and 1.93\% in the non-shadow regions.
Performing the shadow removal by an inpainting model decreases the RMSE 59.46\% in the shadow regions and 32.48\% in the non-shadow regions.


\subsection{Relationship to SOTA Image Inpainting}

\begin{figure}[t]
\centering
\includegraphics[width=1.0\linewidth]{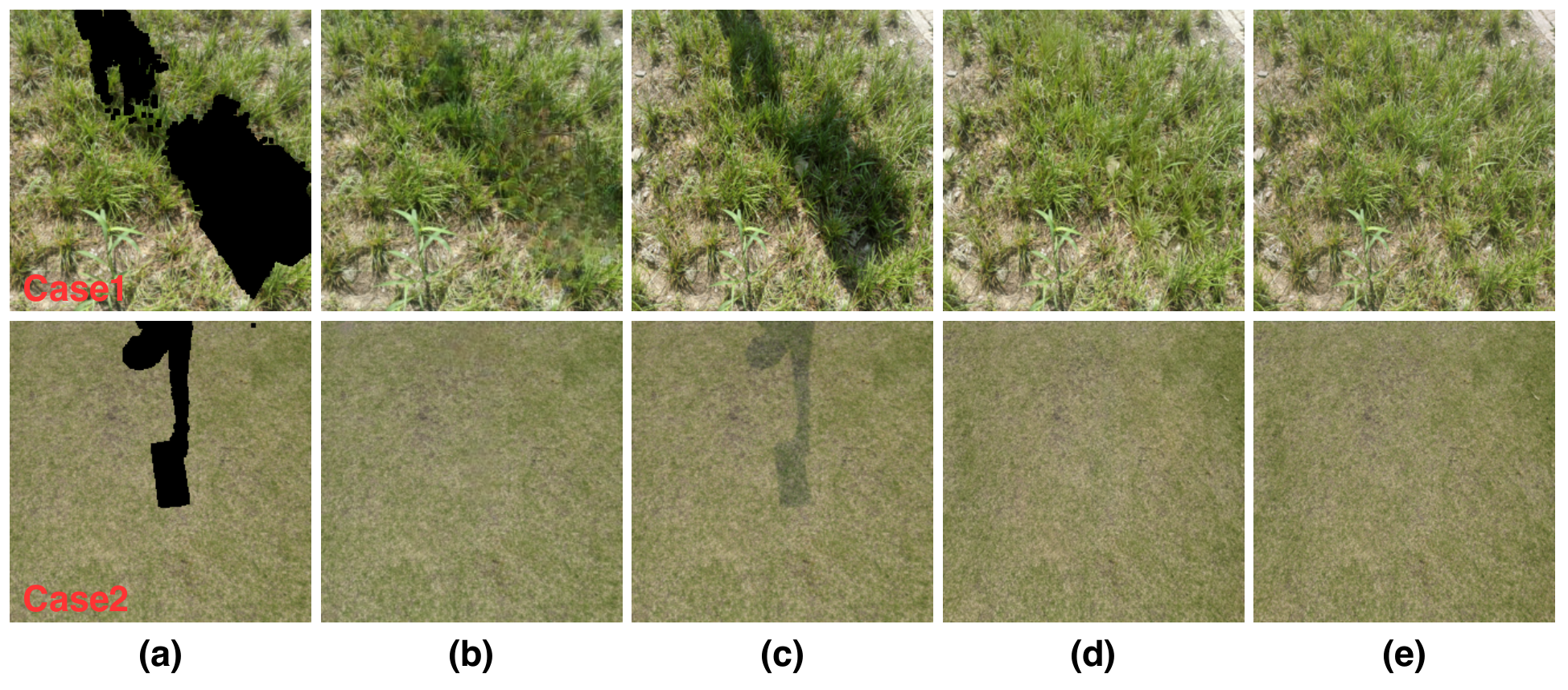}
\vspace{-15pt}
\caption{(a) and (b) are cropped images and restored results of \cite{li2022misf}. (c) and (d) are shadow images and restored results of our method. (e) is the corresponding ground truth.
}
\vspace{-15pt}
\label{fig:comp_misf}
\end{figure}
For the single-image inpainting task, the input is a corrupted image as shown in \figref{fig:comp_misf} (a) which can be got by multiplying the original image with a binary mask. The masked regions are generated totally based on the unmasked background. For the shadow
removal task, the input is a shadow image as shown in \figref{fig:comp_misf} (c) and the shadow regions is indicated by a binary mask. When the binary mask is small or thin \ie (a) and (c) in Case2, the restored result of inpainting is indistinguishable compared with the restored result of shadow removal (see (b) and (d) in Case2). However, when the binary mask is large \ie (a) and (c) in Case1, the texture information of the restored result based on inpainting is worse than the shadow removal-based counterpart. 
This is because the shadow regions \ie (c) can provide more detailed texture information than the masked regions \ie (a). But from (b) in Case1, we can still find that the restored color in the masked regions is similar to the background. This phenomenon proves that the inpainting method has the capability to transfer the color from the background to the masked regions even when the masked regions are large. Based on the above observations, we can find the potential mutual promotion between image inpainting and shadow removal \ie image inpainting and shadow removal can provide color and textual information respectively. 

\subsection{Effectiveness of the inpainting branch and the shadow removal branch}
\begin{figure}[t]
\centering
\includegraphics[width=1.0\linewidth]{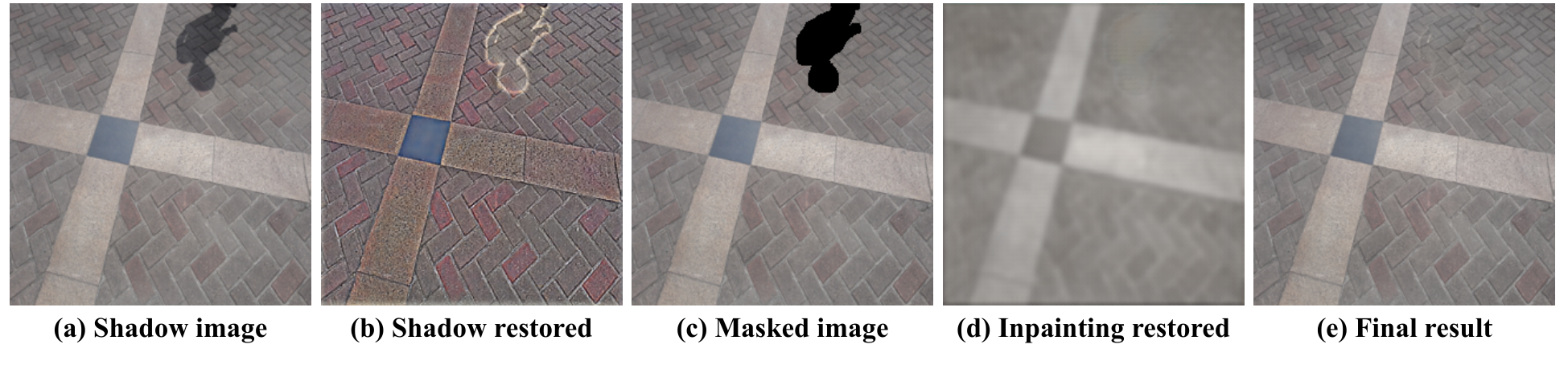}
\vspace{-15pt}
\caption{
From left to right: shadow image, restored result by shadow removal branch, masked image, restored result by inpainting branch, restored result by two branches.
}
\label{fig:one}
\end{figure}

Based on our network architecture (\figref{fig:framework}), we zero out the outputs of the first shadow removal encoder while preserving the second encoder and the decoder for the inpainting task. Remarkably, without any retraining or fine-tuning, the inpainting branch successfully fills in missing pixels (See \figref{fig:one}(c) vs.(d)) with consistent and smooth colors across the affected regions. Similarly, we can remove the inpainting branch to focus solely on shadow removal. As depicted in \figref{fig:one}(a) vs. (b), the colors in the shadow region are recovered and resemble those of shadow-free regions in (a), albeit with some noticeable boundary effects.

%
\subsection{ Effects of inpainting datasets.}
We pretrain our network on three inpainting datasets (See \tableref{tab:dataset}), respectively, and fine-tune on ISTD+ dataset.
In Table~\ref{tab:dataset}, we see that: \ding{182} With different datasets, inpainting pretraining yields significant improvements in shadow removal performance.  \ding{183} Pretraining with a larger dataset leads to better results.
%
\begin{table}[t]
\footnotesize
\centering
\caption{Pretraining our network on three datasets with 10,000 images randomly sampled from each dataset. For the last row, we use full Place2 for pretraining.}
\small
{
    \resizebox{1.0\linewidth}{!}{
    {
	\begin{tabular}{l|cccc|c|c}
		
    \toprule
    
     \multirow{1}{*}{Method} & \multicolumn{4}{c|}{All} & \multicolumn{1}{c|}{Shadow} & \multicolumn{1}{c}{Non-Shadow}   \\
      
    & RMSE$\downarrow$ & PSNR$\uparrow$ & SSIM$\uparrow$ & LPIPS$\downarrow$
    & RMSE$\downarrow$
    & RMSE$\downarrow$  
    \\
    
    \midrule
    w/o pretrain
    & 3.860	& 33.30	& 0.9555 & 0.0599
    & 7.144	
    & 3.263	
    \\
    
    CelabA ($10^4$ images)
    & 3.557	& 33.56	& 0.9586 & 0.0559
    & 6.343 
    & 3.012
     \\
				
    Paris-Street-View ($10^4$ images)
    & 3.511 & 33.64 & 0.9586 & 0.0551
    & 6.286 
    & 2.967 
     \\
					
    Places2 ($10^4$ images)
    & 3.476	& 33.76	& 0.9601 & 0.0550
    & 6.232	
    & 2.937	
     \\

    \topone{Places2 ($1.8\times 10^6$ images)} 
    & \topone{3.398} & \topone{34.14} & \topone{0.9606} & \topone{0.0542} 
    & \topone{5.935}
    & \topone{2.902}
    \\

    \bottomrule

	\end{tabular}
	}
	}
}
\label{tab:dataset}
\vspace{-20pt}
\end{table}

\section{Conclusions}

In this work, we found that pretraining a network on the image inpainting dataset and fine-tuning it on the paired shadow~\&~shadow-free images can enhance the network's capability for shadow removal significantly. We conducted extensive experiments to analyze and compare the networks with or without inpainting pretraining. Inspired by these studies, we formulated shadow removal as an adaptive fusion task and propose a novel adaptive fusion network for shadow removal. The extensive experiments show that our method empowered with inpainting outperforms all state-of-the-art methods. 
%

\paragraph{Acknowledgement.}  This work is supported by the A*STAR Centre for Frontier AI Research and the National Research Foundation, Singapore, and DSO National Laboratories under the AI Singapore Programme (AISG Award No: AISG2-GC-2023-008).

{\small
\bibliographystyle{ieee_fullname}
\bibliography{egbib}
}


\renewcommand\thesection{\Alph{section}}
\renewcommand\thefigure{\Alph{figure}}  
\renewcommand\thetable{\Alph{table}}
\setcounter{section}{0}
\setcounter{figure}{0}

\clearpage
\begin{center}
    \section*{Supplementary Material}
\end{center}

In this material, we add more ablation experiments as well as the detailed architectures of the encoder-decoder mentioned in Sec.\textcolor{red}{3.1} and the network of our method mentioned in Sec.\textcolor{red}{4.2}. 
Besides, we also add a detailed description of the loss functions mentioned in Sec.\textcolor{red}{3.1} and Sec.\textcolor{red}{3.2}. 
To prove the advantage of our methods, we provide some visualized comparison on the image level as well as the $L_1$ difference between the restored result and the corresponding ground truth on the A\&B channels of LAB color space.

\section{More Ablation Experiments}
To further analyze the effectiveness of each part of our proposed method, here we consider another three variants \ie, without sigmoid operation, without $\mathbf{W}_1$ (only use $\mathbf{F}$ without multiplied with $\mathbf{W}_1$), and without $\mathbf{W}_2$ (only use $\tilde{\mathbf{F}}$ without multiplied with $\mathbf{W}_2$). As shown in \figref{tab:ablation_supp}, we can observe that: When removing any of these three variants, the performance decreases over all the metrics.
For the whole image, when removing the sigmoid operation, it increases RMSE 5.86\% and LPIPS 17.53\% as well as decreases PSNR 0.88\% and SSIM 0.50\%.
When removing the $\mathbf{W}_1$, it increases RMSE 1.00\% and LPIPS 3.87\% as well as decreases PSNR 1.14\% and SSIM 0.11\%. 
When removing the $\mathbf{W}_2$, it increases RMSE 2.21\% and LPIPS 3.32\% as well as decreases PSNR 1.03\% and SSIM 0.07\%. 

\begin{table*}[t]
\footnotesize
\centering
\caption{Ablation study on ISTD+ dataset\cite{le2019shadow}.}
\small
\small
{
    \resizebox{0.9\linewidth}{!}{
    {
	\begin{tabular}{l|cccc|ccc|ccc}
		
    \toprule
    
     \multirow{2}{*}{Method} & \multicolumn{4}{c|}{All} & \multicolumn{3}{c|}{Shadow} & \multicolumn{3}{c}{Non-Shadow}   \\
      
    & RMSE$\downarrow$ & PSNR$\uparrow$ & SSIM$\uparrow$ & LPIPS$\downarrow$
    & RMSE$\downarrow$ & PSNR$\uparrow$ & SSIM$\uparrow$
    & RMSE$\downarrow$ & PSNR$\uparrow$ & SSIM$\uparrow$
    \\
    
    \midrule
							
    w/o sigmoid
    & 3.597 & 33.84 & 0.9558 & 0.0637 
    & 7.056 & 37.87 & 0.9864
    & 2.920 & 37.16 & 0.9764
    \\

    w/o W1
    & 3.432 & 33.75 & 0.9595 & 0.0563 
    & 6.065 & 37.96 & 0.9889
    & 2.917 & 36.96 & 0.9769
    \\

    w/o W2
    & 3.473	& 33.79	& 0.9599 & 0.0560 
    & 6.165	& 38.07	& 0.9890
    & 2.946	& 37.03	& 0.9770
    \\
					
    \topone{Ours} 
    & \topone{3.398} & \topone{34.14} & \topone{0.9606} & \topone{0.0542} 
    & \topone{5.935} & \topone{38.46} & \topone{0.9894}
    & \topone{2.902} & \topone{37.27} & \topone{0.9772}
    \\

    \bottomrule

	\end{tabular}
	}
	}
}
\label{tab:ablation_supp}
\vspace{0pt}
\end{table*}

\section{Network Architecture}
\label{subsec:network}

{\bf The Network for Image Inpainting (Sec.\textcolor{red}{3.1})}.We employ a naive encoder-decoder network during the training for image inpainting. The detailed architecture of the encoder-decoder is shown in \tableref{tab:en-decoder}.

\begin{table*}[t]
\setlength{\tabcolsep}{4pt}
\footnotesize
\centering
\caption{Detailed architecture of the naive encoder-decoder network. The input size of each layer is defined as H $\times$ W. The parameters of 'Conv\&ConvTran' are the numbers of input and output channels, kernel size, stride, and padding respectively.}
\small
{
    \resizebox{0.9\linewidth}{!}{
\begin{tabular}{c||l|l||l|l}
\toprule
\multirow{4}{*}{Encoder}
& \multicolumn{2}{l|}{Input size of each layer}          & \multicolumn{2}{l}{Conv. layers}     \\
& \multicolumn{2}{l|}{$256\times256$}    & \multicolumn{2}{l}{Conv(4, 64, 7, 1, 3), ReLU}       \\
& \multicolumn{2}{l|}{$256\times256$}    & \multicolumn{2}{l}{Conv(64, 128, 4, 2, 1), ReLU}     \\
& \multicolumn{2}{l|}{$128\times128$}      & \multicolumn{2}{l}{Conv(128, 256, 4, 2, 1), ReLU}    \\
\midrule
\midrule
\multirow{5}{*}{Decoder} 
&\multicolumn{2}{c||}{Resnet Block $\times$ 8}                & \multicolumn{2}{c}{}                     \\
& Input size of each layer          & Conv. layers                     & Input size of each layer         & Conv. \& ConvTran layers \\
& $64\times64$    & Conv(256, 256, 3, 1, 1),ReLU       & $64\times64$      & ConvTran(256, 128, 4, 2, 1), ReLU \\
& $64\times64$    & Conv(256, 256, 3, 1, 1),ReLU       & $128\times128$    & ConvTran(128, 64, 4, 2, 1), ReLU \\
&                   &                                    & $256\times256$    & Conv(64, 3, 7, 1, 3), ReLU \\
\bottomrule
\end{tabular}
}
}
\vspace{0pt}
\label{tab:en-decoder}
\end{table*}


{\bf The Network of our method (Sec.\textcolor{red}{4.2})}. 
Our adaptive fusion network consists of two encoders, an adaptive fusion block, and a decoder. The detailed architecture of our method is shown in \tableref{tab:ours}. 
%

\begin{table*}[t]
\setlength{\tabcolsep}{4pt}
\footnotesize
\centering
\caption{Detailed architecture of our method. The input size of each layer is defined as H $\times$ W. The parameters of 'Conv\&ConvTran' are the numbers of input and output channels, kernel size, stride, and padding respectively.}
\small
{
    \resizebox{0.9\linewidth}{!}{
\begin{tabular}{c||l|l||l|l}
\toprule
\multirow{5}{*}{Encoder}
&\multicolumn{2}{c||}{Shadow-masked image ($\phi(\cdot)$)}                & \multicolumn{2}{c}{Shadow image ($\psi(\cdot)$)}                     \\
& Input size of each layer          & Conv. layers                     & Input size of each layer         & Conv. layers                    \\
& $256\times256$    & Conv(4, 64, 7, 1, 3), ReLU       & $256\times256$    & Conv(4, 64, 7, 1, 3), ReLU \\
& $256\times256$    & Conv(64, 128, 4, 2, 1), ReLU     & $256\times256$    & Conv(64, 128, 4, 2, 1), ReLU \\
& $128\times128$      & Conv(128, 256, 4, 2, 1), ReLU    & $128\times128$      & Conv(128, 256, 4, 2, 1), ReLU \\
\midrule
\midrule
\multirow{4}{*}{Adaptive Fusion Block} 

&\multicolumn{4}{l}{Conv(512, 512, 3, 1, 1)}      \\   
&\multicolumn{4}{l}{Sigmoid}                     \\     
&\multicolumn{4}{l}{Element-wise multiplication} \\
&\multicolumn{4}{l}{Conv(512, 256, 3, 1, 1)}      \\ 
\midrule
\midrule
\multirow{5}{*}{Decoder} 
&\multicolumn{2}{c||}{Resnet Block $\times$ 8}                & \multicolumn{2}{c}{}                     \\
& Input size of each layer          & Conv. layers                     & Input size of each layer         & Conv. \& ConvTran layers \\
& $64\times64$    & Conv(256, 256, 3, 1, 1),ReLU       & $64\times64$      & ConvTran(256, 128, 4, 2, 1), ReLU \\
& $64\times64$    & Conv(256, 256, 3, 1, 1),ReLU       & $128\times128$    & ConvTran(128, 64, 4, 2, 1), ReLU \\
&                   &                                    & $256\times256$    & Conv(64, 3, 7, 1, 3), ReLU \\
\bottomrule
\end{tabular}
}
}
\vspace{0pt}
\label{tab:ours}
\end{table*}


\section{Loss Functions}
\label{subsec:loss}

{\bf Training Network for Image Inpainting (Sec.\textcolor{red}{3.1})}.
Given a corrupted image $\mathbf{I}$ where the corrupted regions are indicated by a binary mask $\mathbf{M}$
as well as the corresponding ground truth $\mathbf{I}^*$ and restored result $\hat{\mathbf{I}}$ by generator $\mathcal{G}$. 
We follow \cite{li2022misf} and \cite{nazeri2019edgeconnect} employ $\ell_1$ loss, GAN loss, Perceptual loss, and Style loss during the training. Specifically, the $\ell_1$ loss is defined as:
\begin{align}\label{eq:loss_shadow}
\mathcal{L}_{\ell_1} = \frac{1}{\text{Mean}(\mathbf{M})} || \hat{\mathbf{I}} - \mathbf{I}^*||_1 \text{.}
\end{align}
%
The objective functions to optimize the generator $\mathcal{G}$ and discriminator $\mathcal{D}$ are defined as: 
%
\begin{align}\label{eq:loss_inpainting_1}
 \mathcal{L}_{Gen}(\mathcal{G}) = \mathbb{E}_{\mathbf{I}\sim p(\mathbf{I})}\log[1 - \mathcal{D}(\mathcal{G}(\mathbf{I}))] \text{,}
\end{align}
\begin{align}\label{eq:loss_inpainting_2}
  \mathcal{L}_{Dis}(\mathcal{D}) = \frac{1}{2}\mathbb{E}_{\mathbf{I}\sim p(\mathbf{I})}\log[\mathcal{D}(\mathcal{G}(\mathbf{I}))]
  + \frac{1}{2}\mathbb{E}_{\mathbf{I}^*\sim p(\mathbf{I}^*)}\log[1-\mathcal{D}(\mathbf{I}^*)] \text{.}
\end{align}
The GAN loss is defined as:
\begin{align}\label{eq:loss_inpainting_3}
  \mathcal{L}_{GAN} =  \mathcal{L}_{Gen}(\mathcal{G}) + \mathcal{L}_{Dis}(\mathcal{D}) \text{.}
\end{align}
%
%
To calculate the Perceptual loss, we use the VGG-19 pretrained network $\omega$ to extract the feature from  $\mathbf{I}^*$ and $\hat{\mathbf{I}}$. $\omega_i$ denote the feature comes from each layer of $relu1\_1, relu2\_1, relu3\_1, relu4\_1, relu5\_1$ respectively. The Perceptual loss is defined as:
\begin{align}\label{eq:loss_shadow_4}
\mathcal{L}_{Perc} ~\text{=}~   \mathbb{E}\left[\sum_{i=1}^{} \frac{1}{\mathbf{N}_i} ||\omega_i(\mathbf{I}^*) - \omega_i(\hat{\mathbf{I}})||_1 \right]\text{,}
\end{align}
where we set $\mathbf{N}_i$ to 1.
%
%
%
The same features are used to calculate the Style loss which is shown as:
%
\begin{align}\label{eq:loss_shadow_5}
\mathcal{L}_{Style} = \mathbb{E}\left[||\mathbf{G}(\omega_i(\mathbf{I}^* * \mathbf{M})) - \mathbf{G}(\omega_i(\hat{\mathbf{I}} * \mathbf{M}))||_1 \right] \text{,}
\end{align}
where $\mathbf{G}$ is a Gram matrix constructed from the feature $\omega_i$.
Then, the total loss function is defined as: 
\begin{align}\label{eq:loss_inpainting_6}
\mathcal{L}(\hat{\mathbf{I}}, \mathbf{I}^*) = \lambda_1\mathcal{L}_{\ell_1}+\lambda_2\mathcal{L}_{GAN}+\lambda_3\mathcal{L}_{Perc}+\lambda_4\mathcal{L}_{Style} \text{,}
\end{align}
where we set $\lambda_1=1$, $\lambda_2=0.1$, $\lambda_3=0.1$, and $\lambda_4=250$.

{\bf Training Network for Shadow Removal (Sec.\textcolor{red}{3.2}).} Given a shadow image $\mathbf{I}$ where the shadow regions are indicated by a binary mask $\mathbf{M}$ as well as the corresponding ground truth shadow-free image $\mathbf{I}^*$ and restored result $\hat{\mathbf{I}}$. During the training, we only use $\ell_1$ loss which is shown as:
\begin{align}\label{eq:loss_shadow_7}
\mathcal{L}_{\ell_1} = \frac{\lambda_1}{\text{Mean}(\mathbf{M})}  ||\hat{\mathbf{I}} - \mathbf{I}^*||_1 \text{,}
\end{align}
where we set $\lambda_1=1$.

\section{Visualization Comparison}
\label{subsec:visual}

\begin{figure*}
\centering
\includegraphics[width=1.0\linewidth]{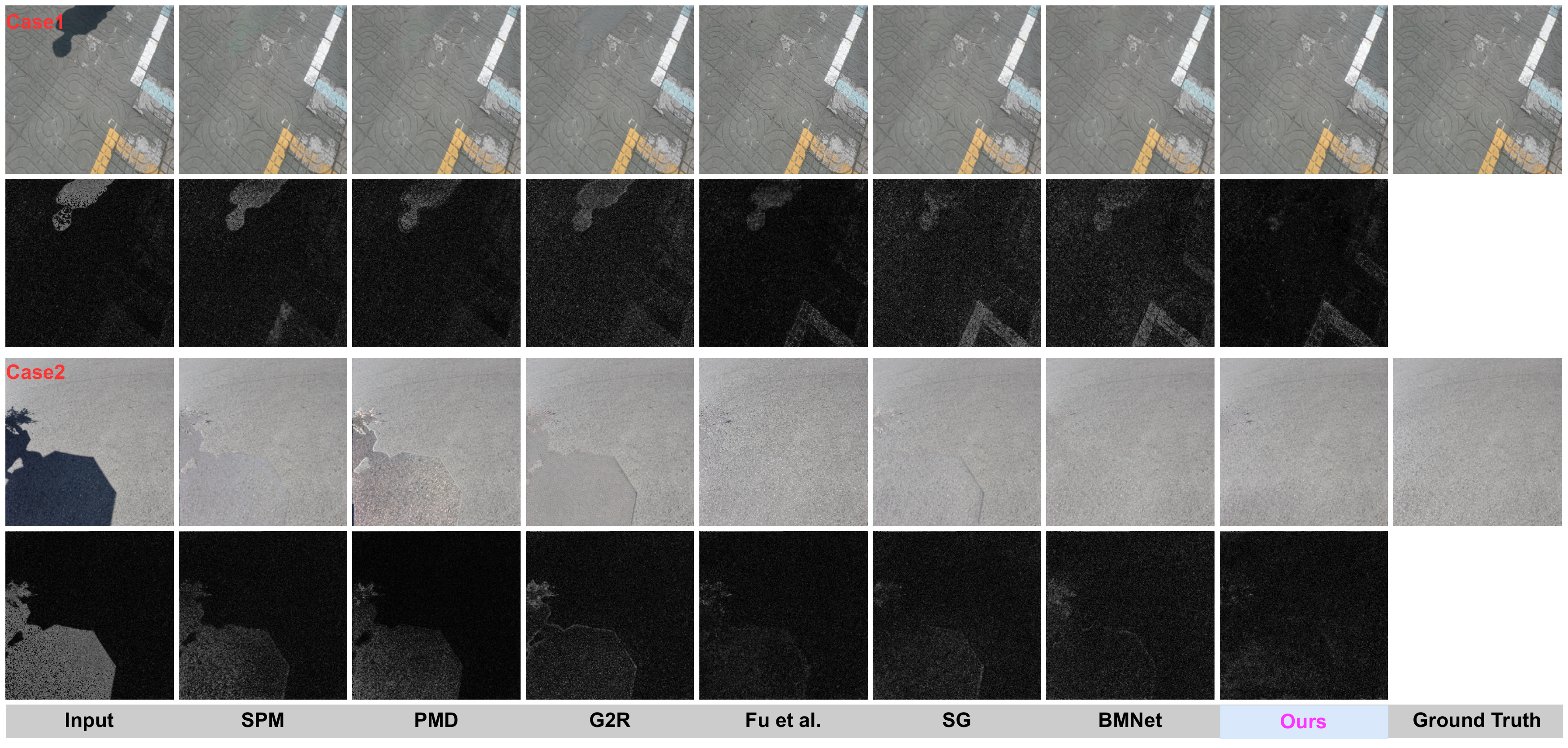}
\vspace{-15pt}
\caption{Comparison on ISTD+ dataset\cite{le2019shadow}. From left to right are input shadow image, SPM\cite{le2019shadow}, PMD\cite{le2020from}, G2R\cite{liu2021from}, Fu et al.\cite{fu2021auto}, SG\cite{wan2022sg-shadow}, BMNet\cite{zhu2022bijective}, Our method, and corresponding ground truth respectively.}

\label{fig:supp_istd+}
\vspace{0pt}
\end{figure*}

\begin{figure*}
\centering
\includegraphics[width=1.0\linewidth]{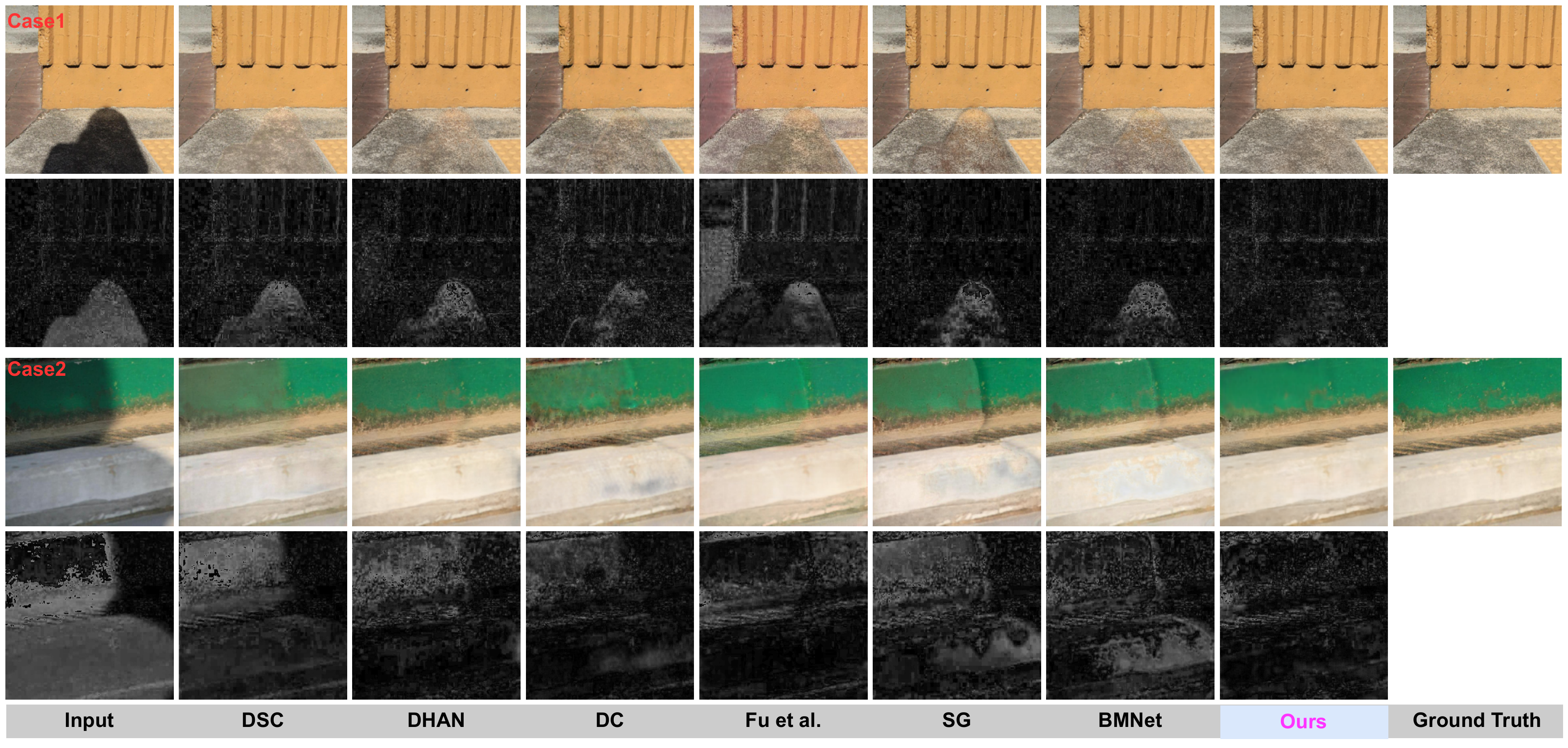}
\vspace{-15pt}
\caption{Comparison on SRD dataset\cite{qu2017deshadownet}. From left to right are input shadow image, DSC\cite{hu2019direction}, DHAN\cite{cun2020towards},  DC\cite{jin2021dc}, Fu et al.\cite{fu2021auto}, SG\cite{wan2022sg-shadow}, BMNet\cite{zhu2022bijective}, Our method, and the corresponding ground truth respectively.
}
\label{fig:supp_srd}
\vspace{0pt}
\end{figure*}

To prove the effectiveness of our method. We have provided the visualized comparison on the image level (see the first and third rows) as well as the $L_1$ difference between the restored result and the corresponding ground truth on A\&B channels of LAB color space (see the second and fourth rows) as shown in \figref{fig:supp_istd+} and \figref{fig:supp_srd}. 
The visualized $L_1$ difference is the averaged results coming from the difference between channels A\&B and their corresponding ground truth. The samples in \figref{fig:supp_istd+} and \figref{fig:supp_srd} are come from ISTD+ dataset\cite{le2019shadow} and SRD dataset\cite{qu2017deshadownet} respectively. 
From the visualization comparison, we can find that 
\ding{182} With our method the shadow boundary in the restored result is more smooth even indistinguishable by human eyes. 
\ding{183} On the LAB color space, the color difference between the restored result and the corresponding ground truth is smaller by using our method.

\end{document}